\pdfoutput=1

\documentclass[11pt]{article}

\usepackage[]{emnlp2021}

\usepackage{times}
\usepackage{latexsym}
\usepackage{graphicx}
\usepackage{booktabs,multicol}
\usepackage[T1]{fontenc}

\usepackage[utf8]{inputenc}

\usepackage{microtype}

\title{Instructions for EMNLP 2021 Proceedings}
\title{Dynamic Forecasting of Conversation Derailment}

\author{Yova Kementchedjhieva \\
  University of Copenhagen \\
  \texttt{yova@di.ku.dk} \\\And
  Anders S{\o}gaard  \\
  University of Copenhagen\\
  \texttt{soegaard@di.ku.dk} \\}

\newcommand{\cga}{\textsc{cga}}
\newcommand{\cmv}{\textsc{cmv}}

\begin{document}
\maketitle
\begin{abstract}


Online conversations can sometimes take a turn for the worse, either due to systematic cultural differences, accidental misunderstandings, or mere malice. Automatically forecasting derailment in public online conversations provides an opportunity to take early action to moderate it. Previous work in this space is limited, and we extend it in several ways. 
We apply a pretrained language encoder to the task, which outperforms earlier approaches. We further experiment with shifting the training paradigm for the task from a static to a dynamic one to increase the forecast horizon. This approach shows mixed results: in a high-quality data setting, a longer average forecast horizon can be achieved at the cost of a small drop in F1; in a low-quality data setting, however, dynamic training propagates the noise and is highly detrimental to performance. 

\end{abstract}

\section{Introduction}

The flow of a conversation can be interrupted when its participants turn to attacking each other as opposed to communicating constructively. The task of forecasting derailment aims to predict outbursts of hostile behavior between users in online conversations \textit{before} they happen \cite{awry}. This stands in opposition to related tasks like hate speech detection \cite{fortuna-etal-2020-toxic}, where abusive language is identified \textit{post factum}. Accurate forecasts could give forum moderators an early warning of troublesome exchanges or even serve as a step towards the prevention, human-driven or automatic, of derailment. Table~\ref{tab:example} shows a conversation that starts off civil and then derails into a personal attack.

\begin{table}
    \centering
    \begin{tabular}{ll}
        A&Where's your source? \\
        B&I'm just looking for it. \\
        A&In other words, it's all bullshit you pulled\\
        &outta your ass, right? \\
    \end{tabular}
    \caption{A conversation derails into a personal attack.}
    \label{tab:example}
\end{table}

As a personal attack can occur at any point during an unfolding conversation, forecasting has to happen dynamically. 
Furthermore, the earlier the warning for a potential derailment comes, the more actionable it is. In the example from Table~\ref{tab:example}, we would therefore attempt a prediction already after the first comment made by user A, and hope to get an early warning for the derailment coming up two turns later. This is indeed the inference procedure that was used in previous work on the task \cite{awry, trouble}. Training, on the other hand, was done statically in those works, i.e. by feeding all turns up to the personal attack as input and making a prediction based on that. This leads to a disparity between the training and inference procedures, where at training time the implicit task of the model is to predict whether the \textit{next} turn contains a personal attack, whereas at inference time its task is to predict whether \textit{any} future turn will contain a personal attack. We propose to bridge this gap by training the model dynamically instead of statically.

We evaluate our dynamic approach on the two available data sets for forecasting derailment, Conversations Gone Awry, built from gold-standard annotations for the task, and Reddit ChangeMyView, built from partial annotations derived from moderator actions. Our experiments reveal that dynamic training does lead to earlier forecasts, but not without a loss in accuracy and F1. In the high-quality data setting the loss is small (0.5\% F1), but in the lower-quality data setting, dynamic training appears to multiply any noise present in the data and is therefore detrimental to model performance. 

A final, somewhat trivial, but necessary contribution of our work is the application of a pre-trained transformer language encoder to the task, which sets a high new baseline, improving over previous results on the two data sets by 2.1 and 1.2 points on the F1 metric, respectively.



\section{Resources for Forecasting Derailment}

In this section, we describe the two main resources available for the task of forecasting derailment.\footnote{We omit PreTox, a fully silver-standard data set, which contains excessive noise \cite{pretox}. }

\paragraph{Conversations Gone Awry (\cga)}
\label{sec:cga} \cga{} \cite{awry} was built from Wikipedia Talk Page conversations. 
The initial seed of conversations is sampled from WikiConv based on an automatic measure of toxicity that ranges from 0 (no toxicity) to 1 \cite{wikiconv}.\footnote{Toxicity scores are obtained with \url{https://www.perspectiveapi.com/}} A conversation is included as a potential example of derailment if the $N$th comment in it has a toxicity score higher than 0.6 and all preceding comments have a score lower than 0.4. A conversation is included as a potential example of non-derailment if all comments in it score below 0.4. This initial seed is subsequently manually annotated (1) to confirm that the initial exchange is civil and (2) to determine whether after this initial exchange a personal attack occurs from one user towards another. Conversations with a personal attack are included in the dataset up to and including the turn containing the personal attack. This means that in effect all $N-1$ turns in a conversation are civil and the $N$th one is optionally civil or contains a personal attack. Topic control is further implemented by pairing each derailed conversation with a non-derailed conversation from the same talk page.
This procedure yields a dataset of 4,188 conversations, split 60-20-20, with a balanced representation of the two classes.

\paragraph{Reddit ChangeMyView (\cmv)}  \citet{trouble} built \cmv{} from Reddit conversations held under the ChangeMyView subreddit, using as partial annotations the deletion of a turn by the platform moderators when that was done under Reddit's Rule 2: "Don't be rude or hostile to other users." Topic control and class balancing are implemented here, too. Yet, there is no control to ensure that all turns in a conversation prior to the last one would be civil, which as the authors point out is a potential source of noise in the data. The resulting dataset consists of 6,842 data points, with a split of 60-20-20.

\section{Methods for Forecasting Derailment}

In this section we describe previous approaches to the task of forecasting derailment in online conversations, as well as the new approach we propose. 

\subsection{Model architectures}
\citet{trouble} encode conversations with a recurrent hierarchical encoder \cite{hred}. In the main model for the task, this encoder is followed by a classification head, making a binary prediction. The authors show that it is beneficial to pretrain the encoder in an encoder-decoder system trained to generate the next turn in a conversation. In their setup, only the first 80 tokens of a turn are considered, presumably for reasons of computational efficiency. 

Recently, pretrained transformer language encoders such as BERT \cite{devlin-etal-2019-bert} have proven successful at various NLP tasks, so we test whether that would extend to forecasting derailment. BERT does not have a hierarchical structure, but relying on attention rather than recurrency it can better capture long-distance dependencies. We preserve information about the boundaries between turns by inserting a [SEP] token between them. One [CLS] token is further added to the start of the input and one [SEP] token to its end. To fit within BERT's maximum sequence length, we crop each conversation down to 512 subword tokens---instead of doing the cropping with a hard limit per turn, however, we crop the longest turn in a conversation by one token recursively, until the overall length is reduced to 512. With 5.2 turns per conversation on average in both \cga{} and \cmv, we can encode 97 subword tokens on average per turn. 

\begin{table*}[th!]
    \centering
    \begin{tabular}{l|ccccc|ccccc}
    \toprule
    &&&\textsc{cga}&&&&&\textsc{cmv}\\
    \midrule
     & Acc & P & R & F1  & Mean H &  Acc & P & R & F1  & Mean H   \\
    \midrule
    CRAFT & 64.4 & 62.7 & 71.7 & 66.9 & 2.36 & 60.5 & 57.5 & 81.3 & 67.3 & 4.01  \\
    BERT$\cdot$SC  & 64.7 & 61.5 & 79.4 &  69.3 & 2.60  & 62.0 & 58.6 & 82.8 &   68.5 &   3.90 \\ 
    BERT$\cdot$SC+  &   64.3 & 61.2 &  78.9 &  68.8 &  2.85 & 56.5 & 56.0 &  73.2 & 61.7 &  4.06 \\ 
    \bottomrule
    \end{tabular}
    \caption{Experimental results. + denotes dynamic training. H denotes the forecast horizon. }
    \label{tab:cga}
\end{table*}

\subsection{Training paradigms}

In this and subsequent sections, we use $N$ to denote the number of turns in a conversation, $f$ to denote the final turn in a conversation (which is always the potential site of derailment in \cga{} and \cmv), $l$ to denote the label of this turn, such that 1=personal attack and 0=civil. We use $t$ to denote any turn in a conversation that is not the final one. A conversation is therefore be formalized as ($\{t_1, ..., t_{N-1}\}, f, l$). 

\paragraph{Static training}\citet{trouble} train their models feeding all $\{t_1, ..., t_{N-1}\}$ turns as input and making a prediction. 
At inference time, the model is tested dynamically, i.e. feeding turn ${t_1}$ as input and making a prediction $\hat{l}_1$, then feeding turns $\{t_1, t_2\}$ and making a prediction $\hat{l}_2$, and so on until $N-1$ predictions have been accumulated. The overall predicted label is obtained as $\hat{l}=\max_{{i=1..N-1}}\hat{l}_i$. \citet{trouble} find that even with static training, early forecasts can be made at inference time---this indicates that derailment is not always a spontaneous phenomenon, but rather one that develops over several turns.

\paragraph{Dynamic training} We hypothesize that dynamic training would allow models to explicitly learn the skill of early forecasting and therefore be better at making forecasts further out, i.e. with a longer horizon. Dynamic training entails that a conversation of N turns ($\{t_1, ..., t_{N-1}\}, f, l$) be mapped into multiple training samples, each representing a different phase of the unfolding conversation, but all labeled the same, i.e. with reference to whether the last turn derails or not.
In the extreme case, we could map a conversation of $N$ turns into $N-1$ training data points: ($\{t_1\}, f, l$), ($\{t_1, t_2\}, f, l$) ... ($\{t_1, ..., t_{N-1}\}, f, l$). As it is not clear how early on in a conversation the signs of derailment can actually be detected, we propose to instead unroll only the last $K$ turns $t$ of a conversation, where $K$ is a tunable hyperparameter. Even so, we acknowledge that turns that are further from the point of derailment are less likely to offer reliable signal for the upcoming attack, so we propose to take a weighted mean over the loss for a minibatch of data points, such that full (up to $N-1$) conversations are weighed by a factor $\alpha$ and  partial conversations are weighed by a factor $\beta$. These weights are also tunable hyperparameters.

\begin{figure*}
    \centering
    \includegraphics[width=\linewidth]{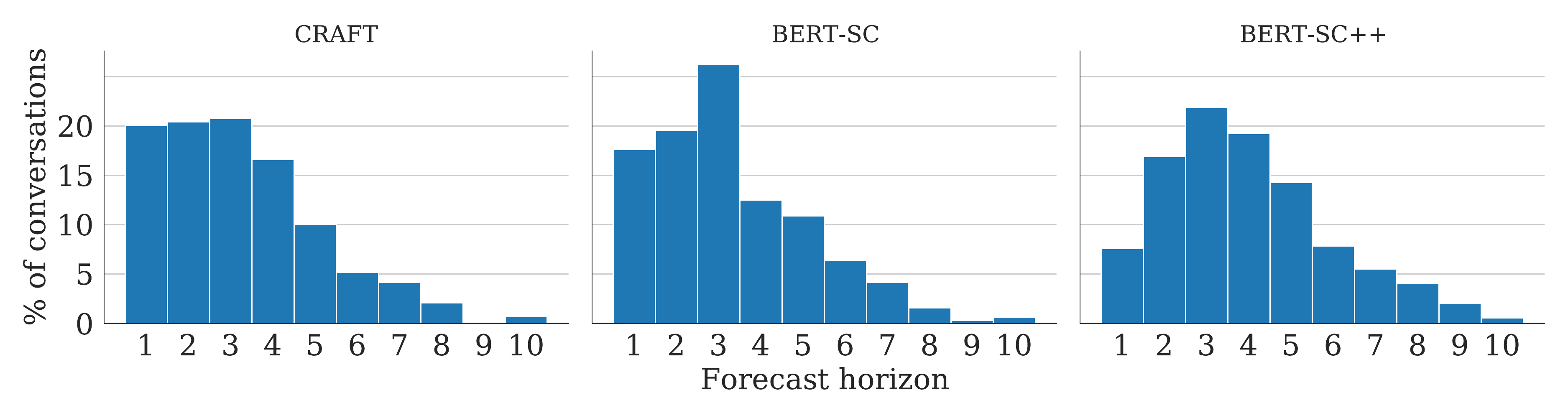}
    \caption{Forecast horizon on the \cga{} data set with a model drawn at random from among the 10 available ones. A horizon of 1 means that an upcoming derailment was only predicted on the last turn before it occurred. }
    \label{fig:cga}
\end{figure*}

\section{Results and Discussion}

In Table~\ref{tab:cga} we report results from various experimental configurations, each averaged over 10 runs with random initialization (of any parameters that have not been pretrained), to account for variance in model performance. Our models were implemented within the HuggingFace library \cite{wolf-etal-2019-term}, using Optuna \cite{akiba2019optuna} to find the best learning rate (from 3e-7 to 1e-5) and batch size (from 16 to 48), based on accuracy on the validation data.\footnote{BERT$\cdot$SC: 6.7e-6 and 32; BERT$\cdot$SC+: 5.4e-6 and 48.} Results are reported in terms of accuracy (Acc), precision (P), recall (R) and F1 score, as well as mean forecast horizon (H). A forecast horizon of 1 means that a derailment coming up on turn $N$ was first detected on turn $N-1$. 

\paragraph{CRAFT v BERT$\cdot$SC} 
CRAFT models were trained using the implementation made available by \citet{trouble}---this is our baseline.\footnote{The results we obtain are lower than what the authors reported, due to the averaging over 10 runs - the best of the 10 runs closely resembles the results reported in the original paper, i.e. the results are reproducible but they also vary.} BERT$\cdot$SC refers to a model consisting of the BERT (\texttt{bert-base-uncased}) checkpoint with a sequence classification (SC) head, trained statically, i.e. in the same manner as CRAFT. We see an improvement in F1 with BERT as compared to CRAFT: 2.1 points for \cga{} and 1.2 for \cmv. BERT$\cdot$SC is slightly less precise, but compensates with a highly improved recall.


\paragraph{BERT$\cdot$SC v BERT$\cdot$SC+}
BERT$\cdot$SC+ refers to a model with the same architecture as BERT$\cdot$SC that was trained dynamically instead of statically. In preliminary experiments we tuned $K$, $\alpha$ and $\beta$ on the development set of \cga. In a comparison of values $3$ and $5$ for $K$ we found the former to give a better F1 score (the latter triggering a high rate of false positives).\footnote{The range of K is determined by the observation that the average length of a conversation in both datasets is just above 5 turns.} We further found the combination of $\alpha=1.5$ and $\beta=0.5$ to work better than $\alpha=1.0, \beta=0.5$. The takeaway here is that dynamic training has to be fairly conservative, presumably because in many cases the eventual derailment of a conversation presents with no early signs. As we see in Table~\ref{tab:cga}, despite the conservative hyperparameter settings, dynamic training results in a longer forecast horizon as intended. BERT$\cdot$SC+ yields slightly lower precision, recall and F1 scores as compared to its statically trained counterpart on the \cga{} dataset---this drop could be considered an acceptable trade-off for the longer forecast window. A much larger drop is seen on these metrics for the \cmv{} dataset, however, rendering the approach unfit for this context. 

\subsection{Forecast horizon: \cga}
Assuming that in some use-cases a small drop in F1 could be an acceptable trade-off for a longer forecast horizon, we perform further analysis of the results for  \cga.  Figure~\ref{fig:cga} sheds more light on the forecast horizon of different models. First we note that despite their static training, both CRAFT and BERT$\cdot$SC make an early forecast (i.e. with a forecast horizon, $H$, longer than 1 turn) at a high rate: only 20.0 percent of CRAFT's forecasts and 17.5 percent of BERT$\cdot$SC's forecasts came on the last turn before the derailment. BERT$\cdot$SC shows an interesting peak at $H=3$, which appears to be a bottleneck of sorts for this model. Turning to BERT$\cdot$SC+ we see that dynamic training has helped in overcoming this bottleneck, shifting a lot of the density from $H<=3$ towards $H>3$. The last-minute forecasts for this model come at a rate of only 7.5 percent. A longer forecast horizon translates into an earlier warning for moderators who might want to delete the upcoming personal attack as soon as it appears on their platform. If one should want to intervene and prevent the derailment from happening, that can also be achieved more easily before the conflict escalates.

\subsection{Noisy data: \cmv}
Dynamic training is far more detrimental to accuracy and F1 score on the \cmv{} data set than it is on the \cga{} data set. This could be attributed to the differing quality of two data sets: \cga{} is a fully-annotated gold standard data set, and \cmv, a partially annotated one, derived from `distant' labels. To estimate the levels of noise in \cmv{} we obtain per-turn toxicity scores for conversations in that data set from \texttt{detoxify} \cite{Detoxify}. 
Using a threshold of 0.5, we discover that 31.9 percent of conversations in \cmv{} contain toxic language in turns prior to the last one. For comparison, in \cga{} this percentage is 0.1---the reason behind that lies in how the data for \cga{} was sampled and subsequently filtered through manual annotations to ensure that all turns up to the last one were civil (see Section~\ref{sec:cga}). The occurrence of toxicity in a conversations prior to the last turn can therefore be seen as noise, and 31.9 percent is a high rate of noise to have in a fairly small data set. 
With dynamic training, a single noisy conversation gets mapped to $K$ training data points, where at least one, but possibly all carry on the noise as well. This would explain the large drop in performance observed with the dynamic training procedure on the \cmv{} dataset.


\section{Conclusion and Final Remarks}
Dynamic training has the desired effect of extending the forecast horizon, at a justifiable cost for high-quality data, but not for noisy data. The horizon of a forecast is important, but a more sophisticated approach may be necessary to achieve a long horizon when data is noisy, e.g. one that considers the exact structure of conversation derailment.

The analysis of the composition of \cga{} and \cmv{} also raises an issue with the definition of the task: it is perhaps better to view the forecasting of \textit{toxic personal attacks} as a subtask within a more general task of forecasting derailment in conversation---with other types of derailment still remaining unexplored. In recent years, a fine-grained typology of antisocial behaviors has been developed \cite{fortuna-etal-2020-toxic}, under which derailment in conversation is seen as the result of various types of antisocial behavior beyond just personal attacks.


\section*{Ethics Statement}

In our paper, we focus on the problem of forecasting conversation derailment. The practical employment of any such system on online platforms has potential positive impact, but several things would be important to first consider, including whether forecasting is fair \cite{pmlr-v97-williamson19a}, how to inform users about the forecasting (in advance, and when the forecasting affects users), and finally what other action is taken when derailment is forecast. See \citet{kiritchenko2020confronting} for a related overview of such considerations, in the context of abusive language detection. 

\section*{Acknowledgments}
This work was funded by Innovations Fund Denmark under the AutoAI4CS	project. 	

\bibliographystyle{acl_natbib}
\bibliography{anthology,acl2021}


\end{document}